\documentclass[11pt,a4paper]{article}
\usepackage[hyperref]{acl2021}
\usepackage{times}
\usepackage{latexsym}
\usepackage{amsfonts}
\usepackage{array}
\usepackage{amsmath}
\usepackage{mathtools}
\usepackage{booktabs}
\usepackage{algorithmicx}
\usepackage{algpseudocode}
\usepackage[ruled,vlined]{algorithm2e}
\usepackage{amssymb}
\usepackage{url}
\usepackage{color}
\usepackage{multirow}
\usepackage{graphicx}
\usepackage{subfigure}
\usepackage{lipsum}
\usepackage{footmisc} 
\usepackage{footnote}
\usepackage{tablefootnote}
\usepackage{afterpage}
\usepackage{adjustbox}
\usepackage[nice]{nicefrac}

\usepackage{microtype}

\aclfinalcopy 
\def\aclpaperid{3882} 

\newcommand*{\affaddr}[1]{#1} 
\newcommand*{\affmark}[1][*]{\textsuperscript{#1}}
\newcommand*{\email}[1]{\texttt{{#1}}}
\usepackage{xstring}

\graphicspath{{figures/}}

\title{Learning Relation Alignment for Calibrated Cross-modal Retrieval}

\author{Shuhuai Ren\affmark[1], \ 
Junyang Lin\affmark[3], \ 
Guangxiang Zhao\affmark[1], \ 
Rui Men\affmark[3], \ 
An Yang\affmark[3], \\ \bf 
Jingren Zhou\affmark[3], \ 
Xu Sun\affmark[1,2]\Thanks{~Corresponding Author}, \ 
Hongxia Yang\affmark[3]\\
\affaddr{\affmark[1]MOE Key Lab of Computational Linguistics, School of EECS, Peking University}\\
\affaddr{\affmark[2]Center for Data Science, Peking University}\\
\affaddr{\affmark[3]Alibaba Group, China}\\
\email{shuhuai\_ren@stu.pku.edu.cn, \{zhaoguangxiang,xusun\}@pku.edu.cn}\\
\email{\{junyang.ljy,menrui.mr,ya235025\}@alibaba-inc.com}\\
\email{\{jingren.zhou,yang.yhx\}@alibaba-inc.com}\\
}

\date{}

\begin{document}
\maketitle
\begin{abstract}
Despite the achievements of large-scale multimodal pre-training approaches, cross-modal retrieval, e.g., image-text retrieval, remains a challenging task. 
To bridge the semantic gap between the two modalities, previous studies mainly focus on word-region alignment at the object level, 
lacking the matching between the linguistic relation among the words and the visual relation among the regions. 
The neglect of such relation consistency impairs the contextualized representation of image-text pairs and hinders the model performance and the interpretability. 
In this paper, we first propose a novel metric, \textbf{I}ntra-modal \textbf{S}elf-attention \textbf{D}istance (\textbf{ISD}), to quantify the relation consistency by measuring the semantic distance between linguistic and visual relations. 
In response, we present \textbf{I}nter-modal \textbf{A}lignment on \textbf{I}ntra-modal \textbf{S}elf-attentions (\textbf{IAIS}), a regularized training method to optimize the ISD and calibrate intra-modal self-attentions from the two modalities mutually via inter-modal alignment.
The IAIS regularizer boosts the performance of prevailing models on Flickr30k and MS COCO datasets by a considerable margin, which demonstrates the superiority of our approach.\footnote{Our code is available at \url{https://github.com/lancopku/IAIS}}
\end{abstract}

\section{Introduction}

\begin{figure}[t!]
    \centering
    \includegraphics[width=\linewidth]{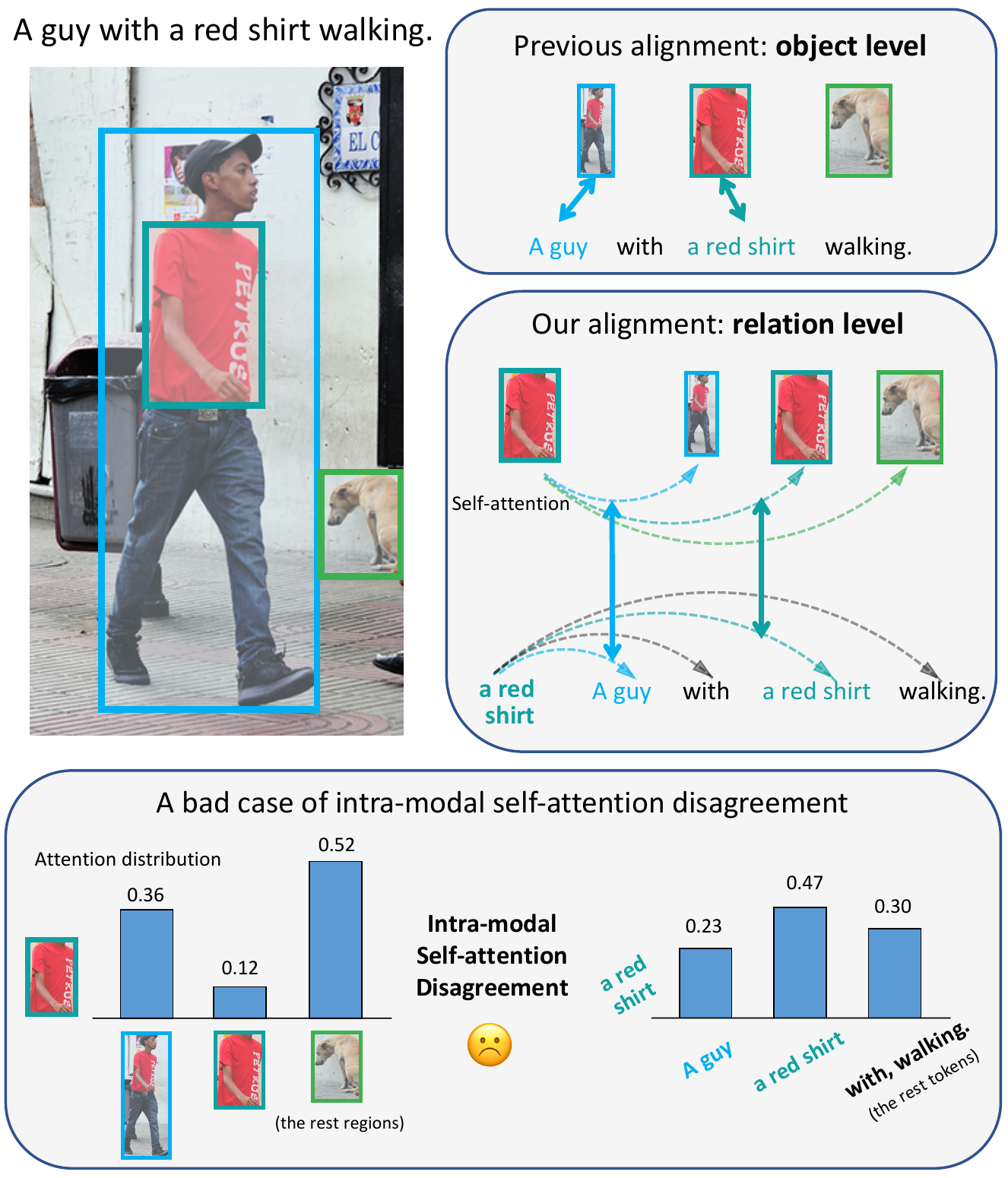}
    \caption{The upper part shows a comparison of previous object-level alignment and our relation-level alignment. The symbol $\leftrightarrow$ denotes alignment and $\dashrightarrow$ denotes the self-attention stems from a query. 
    The lower panel gives a bad case of inconsistent textual and visual relations. The region of ``a red shirt'' pays considerable attention to the region of the dog, which does not benefit the matching and is inconsistent with the self-attention of the corresponding phrase.}
    \label{fig:bad-case}
\end{figure}

Cross-modal retrieval, including image-text retrieval, video-text retrieval, etc., has long been an important downstream task in cross-modal representation learning. 
Image-Text Retrieval (ITR)
aims at modeling the similarity of image-text pairs and recalling the most relevant one.
It remains quite challenging due to the heterogeneity of the data and the semantic gap between two different modalities. 
To bridge this gap, neural networks are responsible for learning global representations of images and texts in a joint semantic space and aligning the images and texts with the same semantics~\citep{VSE++, Unifying-Visual-Semantic}. 
A straightforward way to enhance the alignment is to enforce the local matching between the object-oriented words and the corresponding image regions, and then leverage the object co-occurrence statistics~\cite{Graph-Structured-Network, Denotation-Graph} in the pairs for inference. 
Previous studies incorporate auxiliary knowledge source like scene graphs~\cite{ERNIE-ViL} or object tags~\cite{Oscar} to explicitly indicate the cross-modal mapping.
Other researches try to establish fine-grained interaction on cross-modal attention to reinforce the focus from words to their most relevant regions, and vice versa~\cite{UNITER, CAMP, TERAN, SCAN, DCA, Visual-Agreement}.

However, such word-region alignment at object level serves only as the basis 
because it mainly focuses on the local semantics but lacks the matching of global features like the \textbf{intra-modal relation}. 
The intra-modal relation refers to the correlation of items within a textual or visual sequence. 
More specifically, given a sentence and an image that describe the same scene and are highly matched, the correlation of the items in the textual sequence should also agree with the correlation of the corresponding items in the visual sequence. 
But such constraint of relation consistency is neglected in previous works, which hinders performance and interpretability of the models.
To corroborate this, we conduct a case study on Flickr30k Entities dataset~\cite{Flickr30k-Entities} 
to probe the agreement of relation-level semantics in pre-trained models like UNITER~\cite{UNITER}. 
We utilize the self-attention distribution as a representation of the intra-modal relations~\cite{What-Does-BERT-Look-At, Do-Attention-Heads-Track-Syntactic-Dependencies, Revealing-the-Dark-Secrets-of-BERT}. 

As shown in Figure~\ref{fig:bad-case}, the attention distributions grouped by the annotated object of the given text and image are in disagreement with each other. 
Specifically, the attention distribution in the linguistic modality is reasonable. However, in the visual modality, the region ``a red shirt'' pays inappropriate attention to the region of the dog that doesn't appear in the text, which impairs the representation of this visual item, i.e., ``a red shirt'' under the condition of the corresponding text.
Such mismatched attention distributions suggest that the model represents the same concept with inconsistent semantics, which misleads the model to reduce the estimated similarity of the positive pairs and further leads to the wrong predictions that they are unmatched.
What's even worse is that in practice, the input regions of the existing methods are extracted by a pre-trained object detector like Faster R-CNN~\cite{faster-rcnn}. 
The visual features are much noisier due to over-sampling~\cite{Oscar, BUTD}, which necessitates a stronger regularizer to guide the alignment of the intra-modal relations.

Motivated by the above observations, we promote the semantic alignment from object level to relation level. 
We leverage self-attention matrix to characterize the relation of items within one modality, and design \textbf{I}ntra-modal \textbf{S}elf-attention \textbf{D}istance (\textbf{ISD}), a novel metric to measure 
the consistency between textual and visual relations. 
Our empirical analysis illustrates that the ISD and the model performance on image-text retrieval are highly correlated, which verifies our hypothesis and inspires us to minimize the semantic distance between intra-modal self-attentions in training. 
Accordingly, we propose a new regularized training method called \textbf{I}nter-modal \textbf{A}lignment on \textbf{I}ntra-modal \textbf{S}elf-attentions (\textbf{IAIS}) to calibrate two intra-modal attention distributions mutually via inter-modal alignment, which helps learn better contextualized representations for image-text pairs. 
The model performance of image-text retrieval on Flickr30k and MS COCO datasets is improved by a considerable margin with IAIS, which demonstrates the superiority of our proposal.

\section{Measuring Semantic Distance between Intra-modal Relations}
In this section, we present a formal definition of intra-modal relation alignment (Section~\ref{subsec:formal definition}). 
Such alignment requires extracting the visual and linguistic items corresponding to all objects and sorting them in the same order to make their self-attention distributions comparable. 
We first introduce the mechanism for multimodal attention calculation, and then present the method of attention weight extraction for constructing comparable intra-modal self-attentions (Section~\ref{subsec:Intra-modal Self-attention Reconstruction}).  
Finally, we propose a metric named Intra-modal Self-attention Distance (ISD) to quantify the relation consistency. 
We conduct an empirical analysis on prevailing models to verify the correlation of the model performance and our metric (Section~\ref{subsec:ISD}). 

\subsection{From Intra-modal Relation to Self-attention}
\label{subsec:formal definition}
Given a sequence $\mathbf{O}=[o_1, \cdots, o_N]$ of $N$ objects appeared in an image-text pair, the linguistic and visual representation of such object sequence can be written as $\mathbf{L}=[l_1, \cdots, l_N]$ and $\mathbf{V}=[v_1, \cdots, v_N]$, respectively. 
Each item $l_i, v_i$ with the same index refers to the same object $o_i$.\footnote{An object $o_i$ may require one or more tokens in the text and one or more regions in the image to describe, such that the linguistic item $l_i$ and the visual item $v_i$ may refer to a collection of tokens and regions, respectively.} 
For every object, its relation   
to the others is depicted in both the linguistic and the visual modality. 
From a linguistic view, we regard the following textual self-attention distribution as the relation $\mathbf{R}_{l_i}$ stems from $l_i$: 
\begin{equation}
    \mathbf{R}_{l_i}=[a_{l_i \rightarrow l_1}, \cdots, a_{l_i \rightarrow l_i}, \cdots, a_{l_i \rightarrow l_N}],
\end{equation}
where $a_{l_i \rightarrow l_j}$ is the attention weight from $l_i$ to $l_j$. 
Similarly, the relation $\mathbf{R}_{v_i}$ from the view of the visual modality can be written as
\begin{equation}
    \mathbf{R}_{v_i}=[a_{v_i \rightarrow v_1}, \cdots, a_{v_i \rightarrow v_i}, \cdots, a_{v_i \rightarrow v_N}].
\end{equation}
Consequently, we can achieve relation-level alignment by narrowing the semantic distance, e.g., Kullback-Leibler Divergence, between the linguistic and visual self-attention distribution for all objects from $i=1$ to $N$: 
\begin{equation}
    \min \sum\nolimits_{i=1}^{N} \operatorname{distance} \left( \mathbf{R}_{l_i}, \mathbf{R}_{v_i} \right).
\end{equation}
In the original self-attention matrix, however, the attention weights of specific objects are scattered and disordered. 
We need to extract the target weights and reorder them to construct comparable attention distributions $\mathbf{R}_{l_i}$ and $\mathbf{R}_{v_i}$. 

\subsection{Intra-modal Self-attention Reconstruction}
\label{subsec:Intra-modal Self-attention Reconstruction}
In this subsection, we first introduce the vanilla multimodal attention mechanism and then present a specific way of attention weight extraction. 

Consider models of single-stream Transformer-based architecture like UNITER~\cite{UNITER}. 
The model consists of a stack of Transformer layers with attention mechanism~\cite{Attention-is-all-you-need} and is responsible for encoding image-text pairs into feature representations. 
Given $\mathbf{Q}, \mathbf{K}, \mathbf{V}\in \mathbb{R}^{N \times d}$, 
the matrix of $N$ query, key and value vectors with dimension $d$, respectively, the attention function $\operatorname{Att}(\mathbf{Q}, \mathbf{K}, \mathbf{V})$ is defined as: 
\begin{equation}
\small
    \operatorname{Att}(\mathbf{Q}, \mathbf{K}, \mathbf{V})=\sigma\left(\mathbf{Q} \mathbf{K}^{\top}\right) \mathbf{V}=\sigma\left(\mathbf{S}\right) \mathbf{V}.
\end{equation}
Here, $\sigma$ is a row-wise, scaled softmax and $\mathbf{S}$ is a matrix of attention scores that measure the similarity between every pair of query and key vectors.
Let $\mathcal{L}$ and $\mathcal{V}$ denote the linguistic and the visual modality, respectively. 
Given a textual sequence $\mathbf{X}_{\mathcal{L}}$ of $N_{\mathcal{L}}$ tokens and a visual sequence $\mathbf{X}_{\mathcal{V}}$ of $N_{\mathcal{V}}$ regions, 
the input $\mathbf{X}=\left[\mathbf{X}_{\mathcal{L}} \| \mathbf{X}_{\mathcal{V}}\right]$ in the single-stream architecture is a concatenation of two sequences with length $N=N_{\mathcal{L}}+N_{\mathcal{V}}$. 
\begin{figure}[t]
    \centering
    \includegraphics[width=\linewidth]{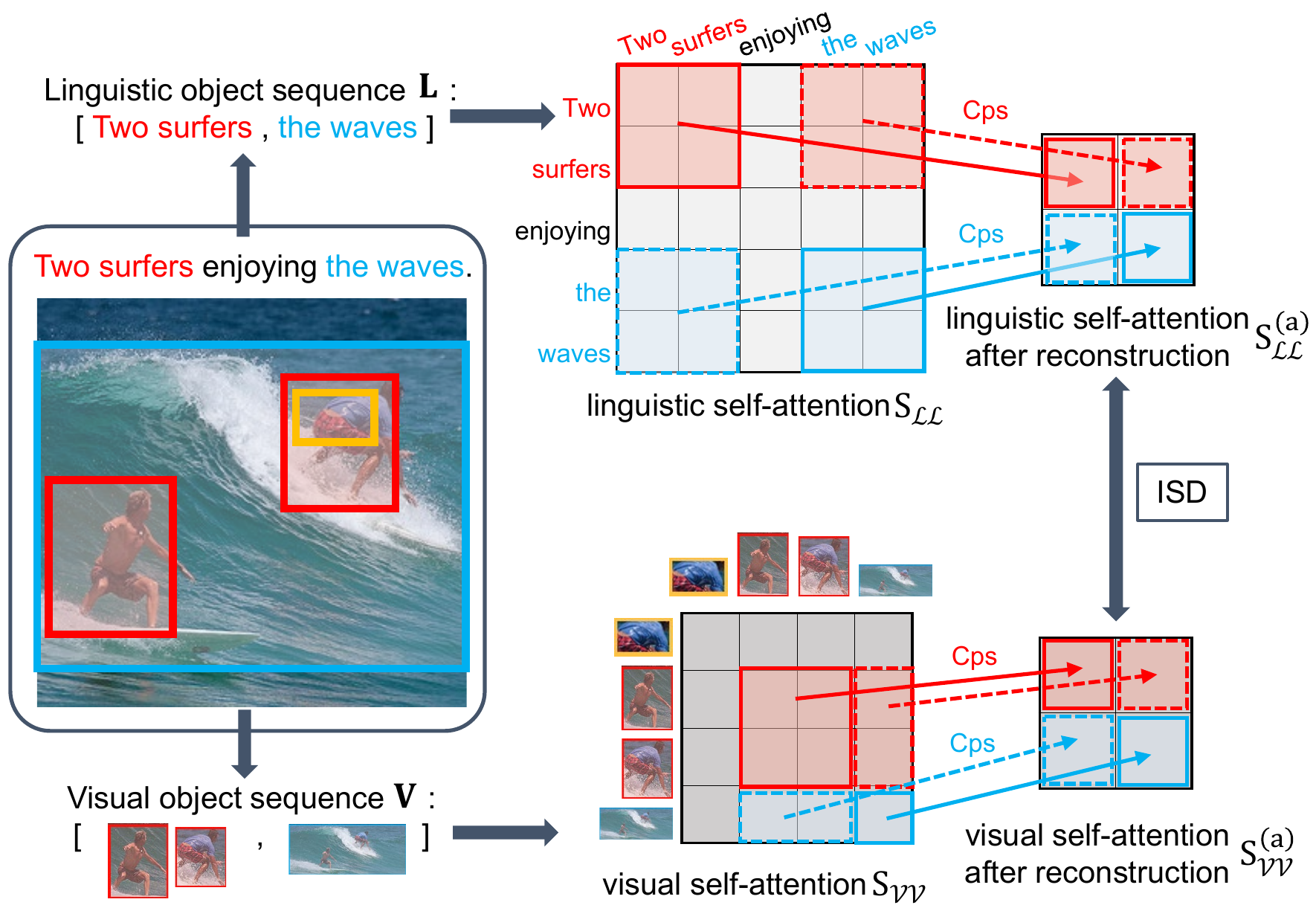}
    \caption{An example of calculating Intra-modal Self-attention Distance (ISDa) for a matched image-text pair. Two inputs in the pair both contain the object of \textit{``two surfers''} and \textit{``the waves''}. For self-attention matrix $\mathbf{S}_{\mathcal{LL}}$ and $\mathbf{S}_{\mathcal{VV}}$ from each modality, we extract object-orientated patches according to the annotations and summarize it with the Cps operation (Eq.~(\ref{equ:cps})) to 
  synthesize new matrices $\mathbf{S}_{\mathcal{LL}}^{(a)}$ and $\mathbf{S}_{\mathcal{VV}}^{(a)}$. Finally, we use our $\operatorname{ISDa}$ metric to measure their semantic distance.}
    \label{fig:msa}
\end{figure}
Accordingly, the query and key matrix\footnote{The value matrix $\mathbf{V}$ is omitted for brevity.} can be written as
\begin{equation}
\small
\begin{split}
    &\mathbf{Q}=\mathbf{X} \mathbf{W}^{Q}=\left(\begin{array}{l}
    \mathbf{X}_{\mathcal{L}} \\
    \mathbf{X}_{\mathcal{V}}
    \end{array}\right) \mathbf{W}^{Q}=\left(\begin{array}{l}
    \mathbf{Q}_{\mathcal{L}} \\
    \mathbf{Q}_{\mathcal{V}}
    \end{array}\right) \\
    &\mathbf{K}=\mathbf{X} \mathbf{W}^{K}=\left(\begin{array}{l}
    \mathbf{X}_{\mathcal{L}} \\
    \mathbf{X}_{\mathcal{V}}
    \end{array}\right) \mathbf{W}^{K}=\left(\begin{array}{l}
    \mathbf{K}_{\mathcal{L}} \\
    \mathbf{K}_{\mathcal{V}}
    \end{array}\right),
\end{split}
\end{equation}
where $\mathbf{W}^{Q}$ and $\mathbf{W}^{K}$ are learnable parameters. 
Furthermore, the attention score matrix $\mathbf{S} \in \mathbb{R}^{N \times N}$ can be organized into four sub-matrices~\cite{Multimodal-Pretraining-Unmasked}:
\begin{equation}
\small
    \begin{aligned}
    \mathbf{S}=\mathbf{Q} \mathbf{K}^{\top} &=\left(\begin{array}{c}
    \mathbf{Q}_{\mathcal{L}} \\
    \mathbf{Q}_{\mathcal{V}}
    \end{array}\right)\left(\mathbf{K}_{\mathcal{L}}^{\top} \mathbf{K}_{\mathcal{V}}^{\top}\right) \\
    &=\left(\begin{array}{ll}
    \mathbf{Q}_{\mathcal{L}} \mathbf{K}_{\mathcal{L}}^{\top} & \mathbf{Q}_{\mathcal{L}} \mathbf{K}_{\mathcal{V}}^{\top} \\
    \mathbf{Q}_{\mathcal{V}} \mathbf{K}_{\mathcal{L}}^{\top} & \mathbf{Q}_{\mathcal{V}} \mathbf{K}_{\mathcal{V}}^{\top}
    \end{array}\right) \\
    &=\left(\begin{array}{ll}
    \mathbf{S}_{\mathcal{LL}} & \mathbf{S}_{\mathcal{L} \mathcal{V}} \\
    \mathbf{S}_{\mathcal{V} \mathcal{L}} & \mathbf{S}_{\mathcal{VV}}
    \end{array}\right).
    \end{aligned}
    \label{equ:sub-matrices}
\end{equation}
The matrices $\mathbf{S}_{\mathcal{LL}}$ and $\mathbf{S}_{\mathcal{VV}}$ on the diagonal represent the linguistic and the visual \textbf{intra-modal} self-attention, respectively. 
$\mathbf{S}_{\mathcal{LV}}$ and $\mathbf{S}_{\mathcal{VL}}$ on back-diagonal represent the \textbf{inter-modal} attention scores from text to image, and the opposite.
We regard the self-attention $\sigma\left(\mathbf{S}_{\mathcal{LL}}\right)$ and $\sigma\left(\mathbf{S}_{\mathcal{VV}}\right)$ as depictions of the intra-modal relations. 
Each row of the matrix represents the relation stemming from one linguistic or visual item to the others within the same modality. 

To construct the comparable intra-modal self-attention matrices, we leverage the object annotations in the Flickr30k Entities dataset~\cite{Flickr30k-Entities} to extract the tokens, regions, and attention weights with respect to the target objects. 
As shown in Figure~\ref{fig:msa}, the text and the image both contain annotated objects of ``two surfers'' and ``the waves''.
The linguistic object sequence can be written as $\mathbf{L}=\left[l_1,l_2 \right]=\left[\text{``two surfers''}, \text{``the waves''}\right]$. 
These two objects derive four intrinsic relations and can be described by four patches in the original linguistic self-attention matrix $\mathbf{S}_{\mathcal{LL}}$.
For clarity, we define an operation $\operatorname{Ext}(\mathbf{S}, o_i, o_j)$ that extracts the patch of attention scores in matrix $\mathbf{S}$ from the object $o_i$ to $o_j$. 
Accordingly, the relation from ``two surfers'' to ``the waves'' can be denoted as $\operatorname{Ext}\left( \mathbf{S}_\mathcal{LL}, l_1, l_2 \right)$. 
To describe the relation with a single value instead of a sub-matrix, we further construct an operation $\text{Cps}(\cdot)$ to summarize the attention patch $\mathbf{S}\in \mathbb{R}^{M\times N}$ to a scalar via column-wise sum and row-wise average:
\begin{equation}
\small
    \text{Cps}(\mathbf{S}) = \left( \sum\nolimits_i^M \sum\nolimits_j^N \mathbf{S}_{ij} \right) / M.
    \label{equ:cps}
\end{equation}
After the above processing, we complete the extraction of the linguistic self-attention $\mathbf{S}_{\mathcal{LL}}$ through grouping the items by annotated object.  
The extraction of visual self-attention $\mathbf{S}_{\mathcal{VV}}$ is similar and the final results are denoted as $\mathbf{S}_{\mathcal{LL}}^{(a)}$ and $\mathbf{S}_{\mathcal{VV}}^{(a)}$.
As our processing for two intra-modal self-attentions follows the same order of object annotations, the matrices $\mathbf{S}_{\mathcal{LL}}^{(a)}$ and $\mathbf{S}_{\mathcal{VV}}^{(a)}$ from two modalities are of the same dimension and comparable.

\subsection{Intra-modal Self-attention Distance with Annotation (ISDa)}
\label{subsec:ISD}
Given two comparable matrices $\mathbf{S}_{\mathcal{LL}}^{(a)}$ and $\mathbf{S}_{\mathcal{VV}}^{(a)}$, we propose a metric called Intra-modal Self-attention Distance with annotation (ISDa) to quantify their semantic gap at the relation level. 
We define the following symmetric matrix-based Kullback-Leibler Divergence ($\operatorname{m-KL}$) for measuring the distance between two matrices $\mathbf{A}$ and $\mathbf{B}$:
\begin{equation}
\small
    \operatorname{m-KL}(\mathbf{A}, \mathbf{B})=\sum\nolimits_i^N \operatorname{KL}\left( \mathbf{A}_i \| \mathbf{B}_i \right) + \operatorname{KL}\left( \mathbf{B}_i \| \mathbf{A}_i \right),
    \label{equ:m-KL}
\end{equation}
where $(\cdot)_i$ stands for the $i^{th}$ row-vector in the matrix and $\operatorname{KL}$ denotes the Kullback-Leibler Divergence. 
Accordingly, the final $\operatorname{ISDa}$ metric for $\mathbf{S}_{\mathcal{LL}}^{(a)}$ and $\mathbf{S}_{\mathcal{VV}}^{(a)}$ is defined as:
\begin{equation}
    \operatorname{ISDa} = \operatorname{m-KL} \Big ( \mathbf{S}_{\mathcal{LL}}^{(a)}, ~ \mathbf{S}_{\mathcal{VV}}^{(a)} \Big ).
\label{equ:ISD}
\end{equation}
We present our algorithm for the calculation of $\operatorname{ISDa}$ in Algorithm~\ref{alg:ISD}.
\newcommand\mycommfont[1]{\small\ttfamily{#1}}
\SetCommentSty{mycommfont}
\SetKwInput{KwInput}{Input}
\begin{algorithm}[t]
\small
\DontPrintSemicolon
\KwInput{Intra-modal self-attention matrices $\mathbf{S}_{\mathcal{LL}}$, $\mathbf{S}_{\mathcal{VV}}$}
\KwInput{Linguistic object sequence $\mathbf{L}$}
\KwInput{Visual object sequence $\mathbf{V}$}
\For{linguistic object $l_i$ in $\mathbf{L}$}{
  \For{linguistic object $l_j$ in $\mathbf{L}$}{
    $\mathbf{S}_{l_i \rightarrow l_j} \leftarrow \operatorname{Ext} \left( \mathbf{S}_{\mathcal{LL}}, l_i, l_j \right)$ \;
    $\mathbf{S}_{\mathcal{LL}}^{(a)}[i,j] \leftarrow \operatorname{Cps} \left(  \mathbf{S}_{l_i \rightarrow l_j} \right)$\;
    }
}
\For{visual object $v_i$ in $\mathbf{V}$}{
  \For{visual object $v_j$ in $\mathbf{V}$}{
    $\mathbf{S}_{v_i \rightarrow v_j} \leftarrow \operatorname{Ext} \left( \mathbf{S}_{\mathcal{VV}}, v_i, v_j \right)$\;
      $\mathbf{S}_{\mathcal{VV}}^{(a)}[i,j] \leftarrow \operatorname{Cps} \left( \mathbf{S}_{v_i \rightarrow v_j}  \right)$\;
    }
}
$\operatorname{ISDa} = \operatorname{m-KL} \Big ( \mathbf{S}_{\mathcal{LL}}^{(a)}, ~ \mathbf{S}_{\mathcal{VV}}^{(a)} \Big )$ \tcp*{Eq.\ref{equ:ISD}}
\Return $\operatorname{ISDa}$
\caption{Intra-modal Self-attention Distance with Annotation (ISDa)}
\label{alg:ISD}
\end{algorithm}

\begin{figure}[t]
    \centering
    \includegraphics[width=0.9\linewidth]{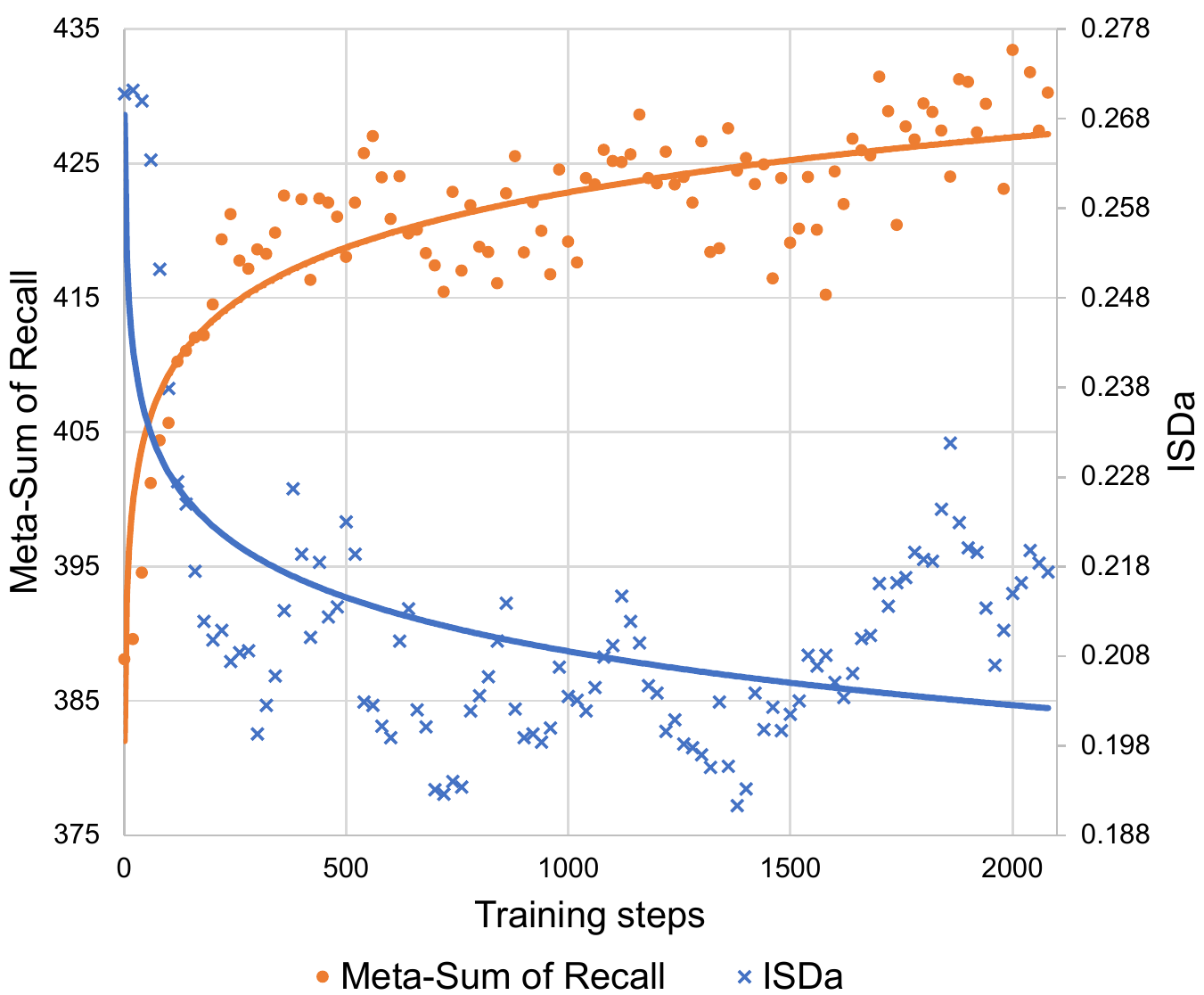}
    \caption{The ISDa (blue $\times$) and model performance (Meta-Sum of Recall, orange $\bullet$) with respect to the training steps. They are highly correlated with a Pearson's correlation coefficient of -0.60.}
    \label{fig:msd-performance}
\end{figure}

\begin{figure*}[t!]
    \centering
    \includegraphics[width=\linewidth]{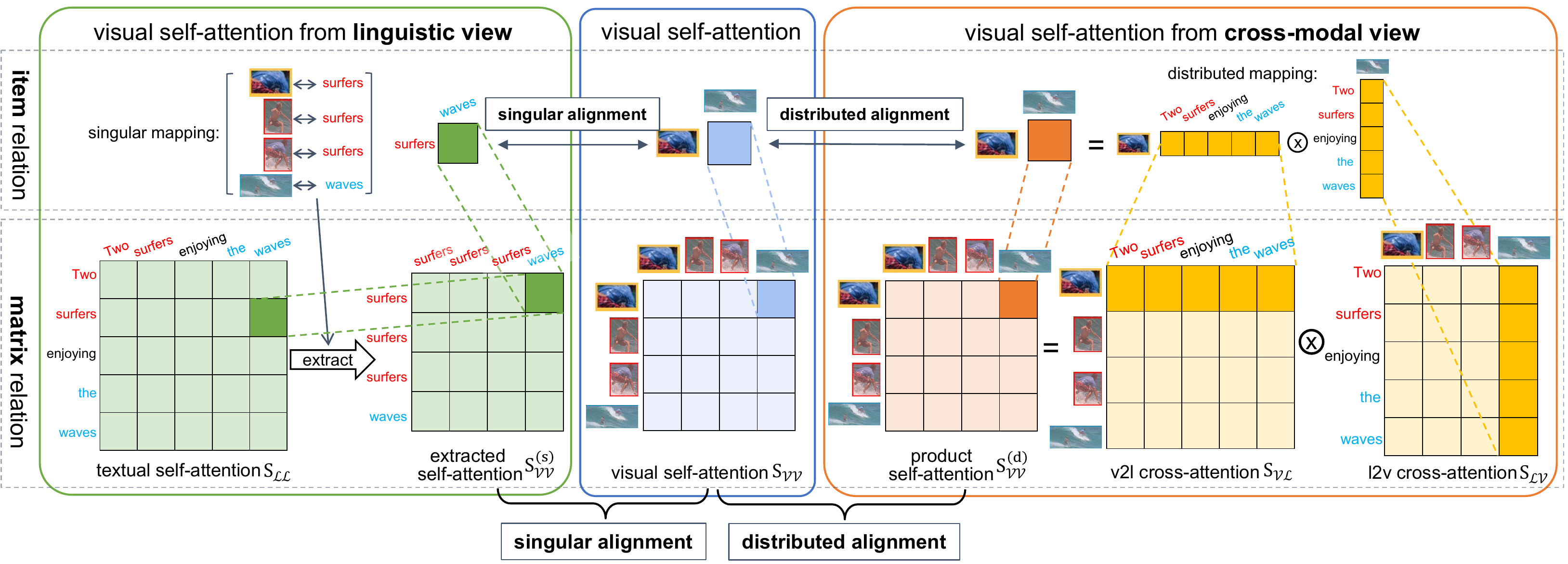}
    \caption{Singular alignment and distributed alignment. The image-text pair here is the same as that in Figure~\ref{fig:msa}.}
    \label{fig:msa-hard-soft}
\end{figure*}

To study the correlation between the ISDa metric and the model performance,\footnote{We use the Meta-Sum~\cite{UNITER}, sum of Recall@1, Recall@5, Recall@10 across the image and text retrieval as a metric for model performance.} we conduct an empirical analysis on UNITER~\cite{UNITER}. 
As shown in Figure~\ref{fig:msd-performance}, the ISDa decreases during the training phase while the model performance continues to increase. 
They are strongly correlated with a Pearson's correlation coefficient of -0.60.
After the middle stage of training, the curve of the model performance and ISDa tends to be flat, suggesting that merely optimizing the task-oriented loss function while neglecting the constraint of relation consistency hinders the model from achieving better performance. 
To eliminate the bottleneck, we can minimize the ISD in the training phase as a regularization to induce further improvement for the ITR task and better the model interpretability.

\section{Inter-modal Alignment on Intra-modal Self-attentions (IAIS)}
In this section, we propose a new regularized training method, Inter-modal Alignment on Intra-modal Self-attentions (IAIS), for image-text retrieval. 
Our goal is to enhance the semantic alignment of relations by minimizing the distance between two intra-modal self-attentions (ISD). 

In practice, given the \textbf{original} visual and linguistic input sequence $\mathbf{V}=[v_1, \cdots, v_{N_\mathcal{V}}]$, $\mathbf{L}=[l_1, \cdots, l_{N_\mathcal{L}}]$ with the scattered items,\footnote{As there are no object annotations in practice, each visual item now refers to only one region. 
Each linguistic item also refers to only one token, even if it is a sub-word.} there are no object annotations and the region features extracted by Faster R-CNN are much noisier~\cite{Oscar, BUTD}, which results in difficulty in grouping the attention weights by ground-truth object. 
The $\operatorname{ISDa}$ thus cannot be used directly as the objective function to minimize. 

To tackle this problem, we regard the input sequence from one modality (e.g., the visual sequence $\mathbf{V}$) as an anchor.
For every item in the anchor sequence, we extract its corresponding representation from the other modality (e.g., one item or a collection of items in the linguistic sequence $\mathbf{L}$) to reconstruct a mirrored sequence. 
After that, the items and their relations within the anchor sequence have a one-to-one correspondence with the items and relations within the mirrored sequence, 
which makes the intra-modal self-attentions derived from the two sequences comparable. 
In the next two subsections, we propose two methods, \textbf{singular alignment} and \textbf{distributed alignment}, to accomplish the attention extraction and reconstruction. 
The former establishes a one-to-one mapping between linguistic and visual attention weight, while the latter establishes a distributed mapping.
Besides, we design two losses $L^{(s)}_\text{IAIS}$ and $L^{(d)}_\text{IAIS}$ as a surrogate of the $\operatorname{ISDa}$ to measure the semantic distance between intra-modal self-attention matrices. 
Finally, we incorporate the surrogate loss minimization as a regularization to calibrate intra-modal self-attentions mutually and achieve the relation-level alignment.

\subsection{Singular Alignment}
For every item in the anchor sequence, singular alignment utilizes the inter-modal attention to find its most relevant item from the opposite modality.  
As the inter-modal attention score quantifies the similarity between the items from two modalities, the visual and the linguistic item with the highest score can be aligned with each other. 
For example, given the $i^{th}$ visual item $v_i$ and the inter-modal attention matrix $\mathbf{S}_{\mathcal{VL}}$, the similarities between $v_i$ and all the linguistic items are depicted in $\mathbf{S}_{\mathcal{VL}}[i,:]$, i.e., the $i^{th}$ row of the matrix. 
Hence the most relevant linguistic item for $v_i$ can be denoted as $l_{i^*}$, where $i^*=\arg\max \mathbf{S}_{\mathcal{VL}}[i,:]$. 
Accordingly, for every weight $a_{v_i \rightarrow v_j}$ in the original visual self-attention matrix $\mathbf{S}_{\mathcal{VV}}$, its corresponding weight $a_{l_{i^*} \rightarrow l_{j^*}}$ in the linguistic self-attention matrix $\mathbf{S}_{\mathcal{LL}}$ can be extracted by the following operation:\footnote{Compared with Section~\ref{subsec:Intra-modal Self-attention Reconstruction}, the $\operatorname{Ext}$ operation here extracts a singular attention weight instead of a patch.} 
\begin{equation}
\small
\begin{split}
     a_{l_{i^*} \rightarrow l_{j^*}} &= \operatorname{Ext}\left( \mathbf{S}_{\mathcal{LL}}, l_{i^*}, l_{j^*} \right), \\
    i^* &=\arg\max \mathbf{S}_{\mathcal{VL}}[i,:], \\
    j^* &=\arg\max \mathbf{S}_{\mathcal{VL}}[j,:],
\end{split}
\end{equation}
as a singular alignment.
After all the extractions, we reconstruct a mirrored matrix $\mathbf{S}_{\mathcal{VV}}^{(s)}$ such that $\mathbf{S}_{\mathcal{VV}}^{(s)}[i,j]=a_{l_{i^*} \rightarrow l_{j^*}}$, which can be regarded as a representation of the original visual self-attention $\mathbf{S}_{\mathcal{VV}}$ from the linguistic view. 
The surrogate loss of ISDa between $\mathbf{S}_{\mathcal{VV}}$ and $\mathbf{S}_{\mathcal{VV}}^{(s)}$ is denoted as $L^{(s)}_\text{IAIS-V}$ when taking vision as the anchor modality. 
The similar processing can also be performed when the linguistic sequence is the anchor. We can generate the matrix $\mathbf{S}_{\mathcal{LL}}^{(s)}$ as a visual representation of the linguistic self-attention $\mathbf{S}_{\mathcal{LL}}$ and define a corresponding loss $L^{(s)}_\text{IAIS-L}$. 

The detailed processing of singular alignment is illustrated in Algorithm~\ref{alg:Hard-Alignment} and Figure~\ref{fig:msa-hard-soft}.
The singular version of IAIS loss is defined as:
\begin{equation}
\small
\begin{split}
    L^{(s)}_\text{IAIS} = & L^{(s)}_\text{IAIS-V} + L^{(s)}_\text{IAIS-L} \\
    = & \operatorname{m-KL} \Big ( \sigma(\mathbf{S}_{\mathcal{VV}})~,~ \sigma(\mathbf{S}_{\mathcal{VV}}^{(s)}) \Big ) + \\
    & \operatorname{m-KL} \Big ( \sigma(\mathbf{S}_{\mathcal{LL}})~,~\sigma(\mathbf{S}_{\mathcal{LL}}^{(s)})  \Big ).
\end{split}
\label{equ:hard-alignment}
\end{equation}
\SetCommentSty{mycommfont}
\SetKwInput{KwInput}{Input}
\begin{algorithm}[t]
\small
\DontPrintSemicolon
\KwInput{Intra-modal self-attention matrices $\mathbf{S}_{\mathcal{LL}}$, $\mathbf{S}_{\mathcal{VV}}$}
\For{$i=1$ \KwTo $N_\mathcal{V}$}{
    $i^* \leftarrow \arg\max \mathbf{S}_{\mathcal{VL}}[i,:]$\;
    \For{$j=1$ \KwTo $N_\mathcal{V}$}{
        $j^* \leftarrow \arg\max \mathbf{S}_{\mathcal{VL}}[j,:]$\;
        $\mathbf{S}_{\mathcal{VV}}^{(s)}[i,j] \leftarrow \operatorname{Ext} \left( \mathbf{S}_{\mathcal{LL}}, l_{i^*}, l_{j^*} \right) $\;
    }
}
\For{$i=1$ \KwTo $N_\mathcal{L}$}{
    $i^* \leftarrow \arg\max \mathbf{S}_{\mathcal{LV}}[i,:]$\;
    \For{$j=1$ \KwTo $N_\mathcal{L}$}{
        $j^* \leftarrow \arg\max \mathbf{S}_{\mathcal{LV}}[j,:]$\;
        $\mathbf{S}_{\mathcal{LL}}^{(s)}[i,j] \leftarrow \operatorname{Ext} \left( \mathbf{S}_{\mathcal{VV}}, v_{i^*}, v_{j^*} \right) $\;
    }
}
$L_\text{IAIS}^{(s)} = \operatorname{m-KL} \Big ( \sigma(\mathbf{S}_{\mathcal{VV}})~,~ \sigma(\mathbf{S}_{\mathcal{VV}}^{(s)}) \Big ) + \operatorname{m-KL} \Big ( \sigma(\mathbf{S}_{\mathcal{LL}})~,~ \sigma(\mathbf{S}_{\mathcal{LL}}^{(s)}) \Big )$\; 
\Return $L_\text{IAIS}^{(s)}$
\caption{Singular Alignment}
\label{alg:Hard-Alignment}
\end{algorithm}

\subsection{Distributed Alignment}
As singular items from different modalities may not be able to give a full representation for each other, we further propose distributed alignment, which utilizes a collection of linguistic items as a representation of a visual item, and vice versa. 
Specifically, given two visual items $v_i$ and $v_j$, we regard the inter-modal attentions $\sigma(\mathbf{S}_{\mathcal{VL}}[i,:])$\footnote{The $i^{th}$ row of $\mathbf{S}_{\mathcal{VL}}$.} from $v_i$ to all linguistic items and $\sigma(\mathbf{S}_{\mathcal{LV}}[:,j])$\footnote{The $j^{th}$ column of $\mathbf{S}_{\mathcal{LV}}$.} from all linguistic items to $v_j$ as a kind of features. 
Hence the original similarity $\mathbf{S}_{\mathcal{VV}}[i,j]=a_{v_i \rightarrow v_j}$ between $v_i$ and $v_j$ can also be modeled as a dot-product of their distributed attention features from the cross-modal view: $\sigma(\mathbf{S}_{\mathcal{VL}}[i,:]) \cdot \sigma(\mathbf{S}_{\mathcal{LV}}[:,j])$.
Such distributed alignment leverages the language as a bridge to draw implicit connections within the visual modality, which can be intuitively regarded as the back-translation~\cite{improving-MT} for multimodal. 
As shown in Figure~\ref{fig:msa-hard-soft}, the distributed version of mirrored self-attention matrix can be constructed by a matrix multiplication of two inter-modal attention matrices:
\begin{equation}
\small
\begin{split}
    & \mathbf{S}_{\mathcal{VV}}^{(d)} = \sigma(\mathbf{S}_{\mathcal{V} \mathcal{L}}) ~ \sigma(\mathbf{S}_{\mathcal{L} \mathcal{V}}), \\
    & \mathbf{S}_{\mathcal{LL}}^{(d)} = \sigma(\mathbf{S}_{\mathcal{L} \mathcal{V}}) ~ \sigma(\mathbf{S}_{\mathcal{V} \mathcal{L}}).
\end{split}
\end{equation}
Similar to the version of singular alignment, the distributed IAIS loss can be written as: 
\begin{equation}
\small
\begin{split}
    L^{(d)}_\text{IAIS} = & L^{(d)}_\text{IAIS-V} + L^{(d)}_\text{IAIS-L} \\ 
    = & \operatorname{m-KL} \Big ( \sigma(\mathbf{S}_{\mathcal{VV}})~,~ \mathbf{S}_{\mathcal{VV}}^{(d)} \Big ) + \\
    & \operatorname{m-KL} \Big ( \sigma(\mathbf{S}_{\mathcal{LL}})~,~ \mathbf{S}_{\mathcal{LL}}^{(d)} \Big ).
\end{split}
\label{equ:soft-alignment}
\end{equation}

\begin{table*}[t]
\begin{adjustbox}{max width=\textwidth}
\begin{tabular}{@{}lcccccccccccccc@{}}
\toprule
 &
  \multicolumn{7}{c}{\textbf{Flickr30k}} &
  \multicolumn{7}{c}{\textbf{MS COCO}} \\ \cmidrule(lr){2-8} \cmidrule(l){9-15}
 &
  \multicolumn{3}{c}{\textbf{Image Retrieval}} &
  \multicolumn{3}{c}{\textbf{Text Retrieval}} &
  \textbf{Overall} &
  \multicolumn{3}{c}{\textbf{Image Retrieval}} &
  \multicolumn{3}{c}{\textbf{Text Retrieval}} &
  \textbf{Overall} \\ \cmidrule(lr){2-4} \cmidrule(lr){5-7} \cmidrule(lr){8-8} \cmidrule(lr){9-11} \cmidrule(lr){12-14} \cmidrule(l){15-15}     
\textbf{Model} &
  R@1 &
  R@5 &
  R@10 &
  R@1 &
  R@5 &
  R@10 &
  Meta-Sum &
  R@1 &
  R@5 &
  R@10 &
  R@1 &
  R@5 &
  R@10 &
  Meta-Sum \\ \midrule
UNITER-base$^*$ &
  72.52 &
  92.36 &
  96.08 &
  85.90 &
  97.10 &
  98.80 &
  542.76 &
  50.33 &
  78.52 &
  87.16 &
  64.40 &
  87.40 &
  93.08 &
  460.89 \\
UNITER-base$^\dagger$ &
  72.70 &
  92.60 &
  96.14 &
  85.50 &
  97.30 &
  98.60 &
  542.84 &
  50.41 &
  78.33 &
  86.94 &
  65.16 &
  87.60 &
  93.14 &
  461.58 \\ \midrule
~+~IAIS-singular &
  73.54 &
  \textbf{93.14} &
  \textbf{96.32} &
  86.10 &
  \textbf{98.10} &
  99.10 &
  546.30 &
  50.99 &
  \textbf{78.85} &
  \textbf{87.41} &
  \textbf{66.98} &
  \textbf{89.10} &
  94.02 &
  \textbf{467.35} \\
~+~IAIS-distributed &
  \textbf{73.66} &
  92.88 &
  96.28 &
  \textbf{87.10} &
  97.90 &
  \textbf{99.20} &
  \textbf{547.02} &
  \textbf{51.10} &
  78.70 &
  87.09 &
  66.88 &
  88.90 &
  \textbf{94.10} &
  466.77 \\ \midrule
UNITER-large$^*$ &
  73.56 &
  \textbf{94.08} &
  \textbf{96.76} &
  87.30 &
  98.00 &
  99.20 &
  548.90 &
  52.93 &
  79.93 &
  87.95 &
  65.68 &
  88.56 &
  93.76 &
  468.81 \\
UNITER-large$^\dagger$ &
  75.98 &
  93.40 &
  96.68 &
  85.80 &
  97.80 &
  98.80 &
  548.46 &
  52.57 &
  79.76 &
  88.00 &
  64.24 &
  88.00 &
  93.62 &
  466.19 \\ \midrule
~+~IAIS-singular &
  \textbf{76.86} &
  93.30 &
  95.72 &
  \textbf{88.30} &
  98.40 &
  \textbf{99.40} &
  \textbf{551.98} &
  53.17 &
  \textbf{80.07} &
  87.94 &
  \textbf{67.78} &
  \textbf{89.70} &
  \textbf{94.48} &
  \textbf{473.14} \\
~+~IAIS-distributed &
  76.28 &
  93.32 &
  95.58 &
  \textbf{88.30} &
  \textbf{98.60} &
  99.30 &
  551.38 &
  \textbf{53.18} &
  79.99 &
  \textbf{88.18} &
  67.68 &
  89.34 &
  94.02 &
  472.39 \\ \bottomrule
\end{tabular}
\end{adjustbox}
\caption{Results of image and text retrieval on Flickr30k and MS COCO. 
R@K corresponds to whether the ground truth is recalled among top K results. 
$*$ denotes the results of UNITER taken from~\citet{UNITER} and {$\dagger$} denotes our reproduction. IAIS-singular and ISA-distributed denote the singular and distributed version of the proposed relation-leve alignment, respectively. }
\label{tb:main-results}
\end{table*}

\subsection{Relation Alignment as Regularizer}
\label{subsec:Relation Alignment as Regularizer}
With the IAIS loss, the surrogate of semantic distance between two intra-modal self-attentions, we present a new regularized training method to enhance the relation alignment for image-text retrieval. 
Our final loss is two-fold. The first is the task-orientated margin loss:
\begin{equation}
\small
    L_\text{margin} = \sum\nolimits_{i=1}^{N_p} \left[ \sum\nolimits_{j=1}^{N_n} S_j - S_i + \alpha \right]_+,
    \label{equ:margin-loss}
\end{equation}
where $[x]_+=\max (0,x)$ and $\alpha$ is a preset margin. 
$N_p$ and $N_n$ denote the number of positive and negative pairs. 
$S_i$ and $S_j$ are the similarity scores of a positive and negative image-text pair, respectively. 
The second is the IAIS loss for all \textbf{positive} pairs that quantifies their relation distance. 
The IAIS loss is computed based on the attentions from the last Transformer-layer, and it can be either the singular alignment version (Eq.~(\ref{equ:hard-alignment})) or the distributed alignment version (Eq.~(\ref{equ:soft-alignment})). 
To summarize, our final final loss can be formalized as: 
\begin{equation}
        L = L_\text{margin} + \lambda_t L_\text{IAIS},
\label{equ:final-loss}
\end{equation}
where $\lambda_t$ is a hyper-parameter w.r.t training steps $t$ to balance two loss items. 
Since our relation-level alignment is based on mappings between linguistic and visual items, it is beneficial to focus on the item-level alignment at the previous training stage via the task-orientated loss. Accordingly, we utilize Training Signal Annealing~\cite{uda} to gradually incorporate the signal of the IAIS loss and design the following exponential schedule:
\begin{equation}
\small
    \lambda_t = \exp\left( \left(\nicefrac{t}{T} - 1\right) \times 5 \right).
\label{equ:exp-schedule}
\end{equation}
Here $T$ is the total training steps during fine-tuning phase and $t$ is the current step. 
As a pluggable regularizer, our IAIS method does NOT incorporate any extra parameters and additional data collection yet empowers the models to capture the higher-level semantics of relation consistency efficiently.

\section{Experimental Settings}
\subsection{Benchmark Datasets}
We conduct experiments on the Flickr30k~\cite{Flickr30k} and MS COCO~\cite{COCO} datasets. 
Flickr30K contains 31K images collected from the Flickr website, with five textual descriptions per image. We follow \citet{Deep-visual-semantic-alignments} to split the data into 30K/1K/1K training/validation/test splits. 
MS COCO consists of 123K images, each accompanied with five human-written captions. 
Following \citet{Deep-visual-semantic-alignments}, the data is divided into 82K/5K/5K training/validation/test images. 

\subsection{Fine-tuning Settings}
Due to the limitation of computing resource, we only incorporate IAIS regularization in the phase of fine-tuning instead of pre-training. 
We use the base (12 layers) and the large (24 layers) version of UNITER~\cite{UNITER}, one of the most prevailing large-scale pre-trained models, as our baseline and backbone for IAIS. 
We follow the fine-tuning setting and hyper-parameter configuration of the original paper.\footnote{\url{https://github.com/ChenRocks/UNITER}}  
The margin in Eq.~(\ref{equ:margin-loss}) is 0.2. 
For each positive instance, 31 hard negative instances are sampled on the text and image side, respectively, and as each batch contains 8 different positive instances, the batch size is 512.
The learning rate is 5e-5 and the training steps are 5000 for both base and large models. 
All experiments are run on 8 NVIDIA V100 GPUs. 

\section{Results and Analysis}
\subsection{Main Results}
The main results of the UNITER performance with and without our IAIS regularization are reported in Table~\ref{tb:main-results}. 
Our methods of both singular and distributed version surpass the baseline by a considerable margin. 
The average improvement over all datasets and models is \textbf{4.49}.

\begin{figure}[t]
    \centering
    \includegraphics[width=\linewidth]{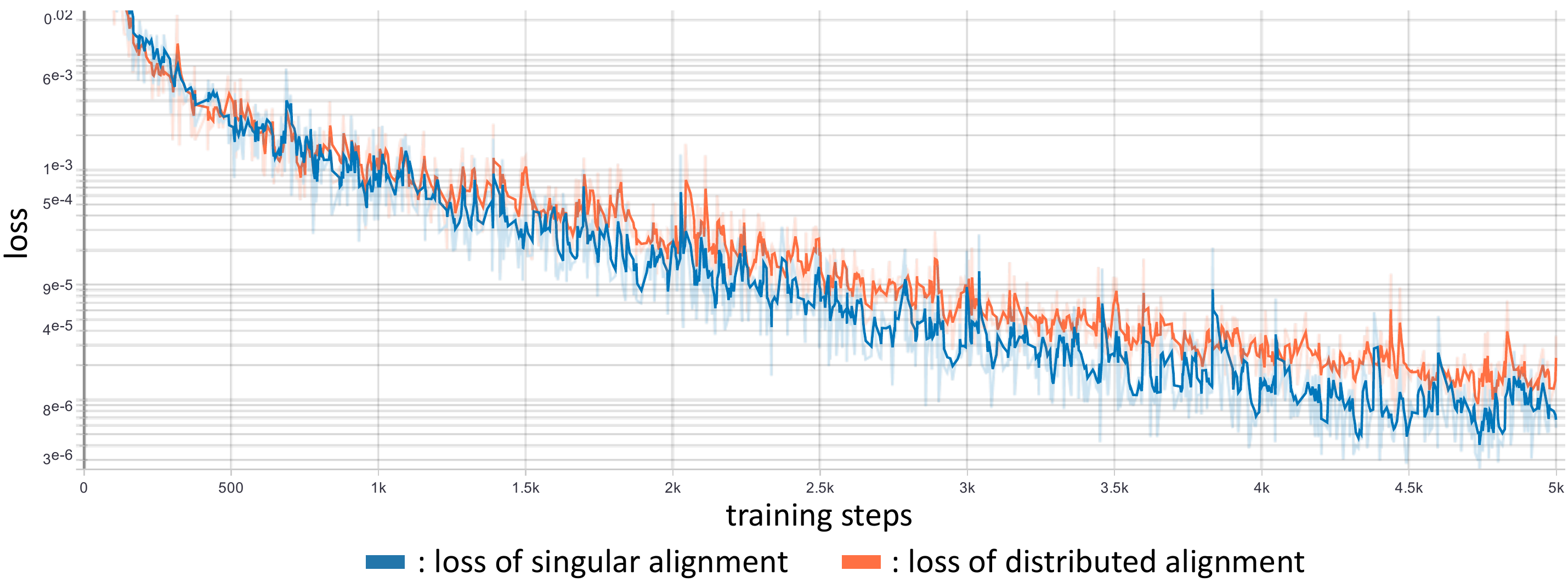}
    \caption{The singular and distributed version of IAIS loss with respect to the training steps.}
    \label{fig:IAIS-loss}
\end{figure}

There are also some interesting findings: 
(1) Compared with image retrieval, the model performance on text retrieval is boosted by IAIS more remarkably with an average improvement of 3.50.
Note that each image in both datasets is paired with five ground-truth sentences, and our IAIS regularizer helps the model capture the common relations for the image and the corresponding texts so that more ground-truth texts can be successfully retrieved.
(2) The improvement on UNITER-base is 17.2\% higher than that on UNITER-large. 
A consistent result can be found in Table~\ref{tb:ISD}, which demonstrates various relation distance metrics of fine-tuned models. 
The ISDa of UNITER-large is smaller than that of UNITER-base, indicating UNITER-large learns more about the relation consistency due to its large capability while there is still room to improve  the relation alignment with our IAIS method. 
(3) 
The relative improvement brought by the singular version of IAIS is 7.0\%, higher than that of the distributed version. 
The ISDa and $L^{(s)}_\text{IAIS}$ are correlated with a Pearson's correlation coefficient of 0.779, which is also higher compared to $L^{(d)}_\text{IAIS}$ with 0.774. 
Besides, our empirical analysis in Figure~\ref{fig:IAIS-loss} shows that it is slightly easier to optimize the $L^{(s)}_\text{IAIS}$, indicating it is a better surrogate of ISDa. 

\begin{table}[t]
\centering
\begin{adjustbox}{max width=0.38\textwidth}
\begin{tabular}{lccc}
\toprule
Model & $\operatorname{ISDa}$          & $L^{(s)}_\text{IAIS}$                        & $L^{(d)}_\text{IAIS}$                        \\ \midrule
UNITER-base    & 0.26          & 0.59                        & 0.36                        \\
~+~IAIS-singular & 0.18          & \textbf{1.31e-3} & \textbf{2.58e-3} \\
~+~IAIS-distributed      & \textbf{0.17} & 2.80e-3 & 2.72e-3 \\ \midrule
UNITER-large   & 0.23          & 0.40                        & 0.16                        \\
~+~IAIS-singular & \textbf{0.18} & \textbf{2.27e-3} & \textbf{3.22e-3} \\
~+~IAIS-distributed      & \textbf{0.18} & 3.15e-3 & 3.70e-3 \\ \bottomrule
\end{tabular}
\end{adjustbox}
\label{ISD}
\caption{Different relation distance metrics of each model after fine-tuning. Lower is better.}
\label{tb:ISD}
\end{table}

\subsection{Effect of Anchor Modality}
In Section~\ref{subsec:Relation Alignment as Regularizer}, we leverage both the linguistic and the visual input as the anchor sequence to reconstruct the mirrored sequence from the opposite modalities. 
To study the impact of the anchor modality, we conduct an ablation study and the results are listed in Table~\ref{tb:ablation-study}. 
Compared to using language as the anchor modality, i.e., only $L_\text{IAIS-L}$ is incorporated, the overall model performance is 2.1 higher when vision is taken as the anchor. 
An explanation is that the description capability of visual regions is more concrete and powerful. 
However, introducing both $L_\text{IAIS-V} + L_\text{IAIS-L}$ to the final loss can achieve a further improvement of 2.22, which indicates the necessity of such combination.

\subsection{Effect of Annealing Schedule}
\begin{figure}[t!]
    \centering
    \includegraphics[width=0.8\linewidth]{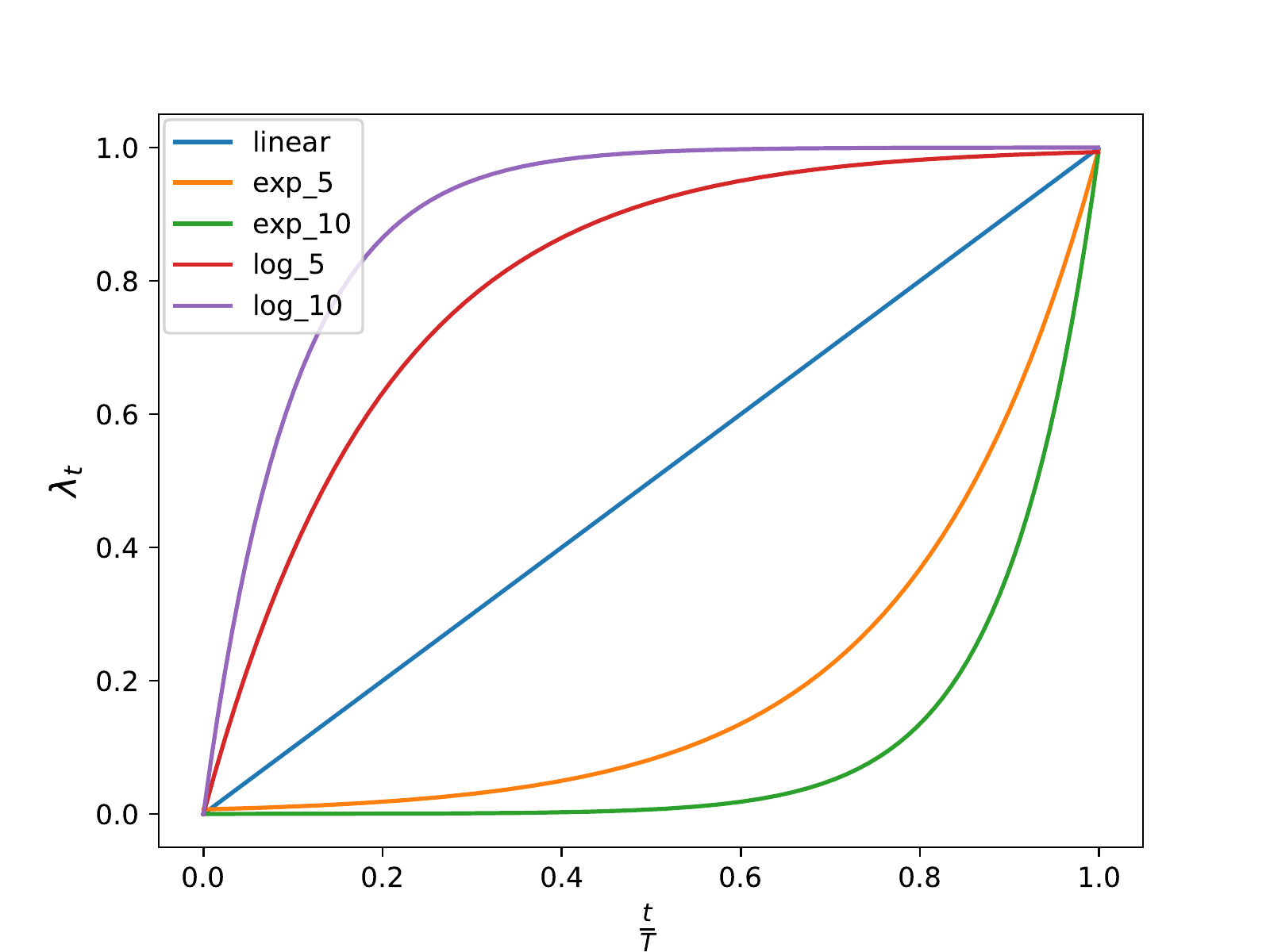}
    \caption{Schedules for IAIS signal annealing.}
    \label{fig:tsa-schedule}
\end{figure}

Besides the exp schedule in Eq.~(\ref{equ:exp-schedule}) for training signal annealing, we also try other schedules: 
\begin{itemize}
    \item log schedule: $\lambda_t = 1-\exp\left(-\nicefrac{t}{T}\times\gamma \right)$;
    \item linear schedule: $\lambda_t = \nicefrac{t}{T}$;
    \item exp schedule: $\lambda_t = \exp\left( \left(\nicefrac{t}{T} - 1\right) \times\gamma \right)$,
\end{itemize}
\noindent where $\gamma$ is chosen from $\{5,10\}$.
All the schedules are shown in Figure~\ref{fig:tsa-schedule}. 
\begin{table}[t!]
\begin{adjustbox}{max width=0.48\textwidth}
\begin{tabular}{lcccccc}
\toprule
               & \multicolumn{3}{c}{\textbf{Image Retrieval}} & \multicolumn{3}{c}{\textbf{Text Retrieval}} \\ \cmidrule(lr){2-4} \cmidrule(lr){5-7} 
\textbf{Model} & R@1   & R@5   & R@10                         & R@1           & R@5          & R@10         \\ \midrule
UNITER-base$^*$    & 72.52 & 92.36 & 96.08                        & 85.90         & 97.10        & 98.80        \\
UNITER-base$^\dagger$    & 72.70 & 92.60 & 96.14 & 85.50         & 97.30        & 98.60        \\ \midrule
~+~IAIS-singular-L    & 72.74 & 92.74 & \textbf{96.12}                        & 86.90         & 96.70        & 99.00        \\
~+~IAIS-singular-V    & \textbf{73.44} & \textbf{92.76} & 95.96                        & \textbf{87.00}  & \textbf{97.50}  & \textbf{99.10}        \\ \midrule
~+~IAIS-distributed-L    & 72.48 & \textbf{92.96} & \textbf{96.26}                        & 86.90         & 97.10        & 99.10        \\
~+~IAIS-distributed-V    & \textbf{73.14} & 92.44 & 96.06                        & \textbf{87.10}  & \textbf{97.40}  & \textbf{99.20}        \\ \bottomrule
\end{tabular}
\end{adjustbox}
\caption{Ablation study on the Flickr30k dataset. ``-L'' denotes that only $L_\text{IAIS-L}$ is incorporated, which regards language as the anchor modality. Similar for ``-V''.}
\label{tb:ablation-study}
\end{table}

\begin{figure}[t!]
    \centering
    \includegraphics[width=0.9\linewidth]{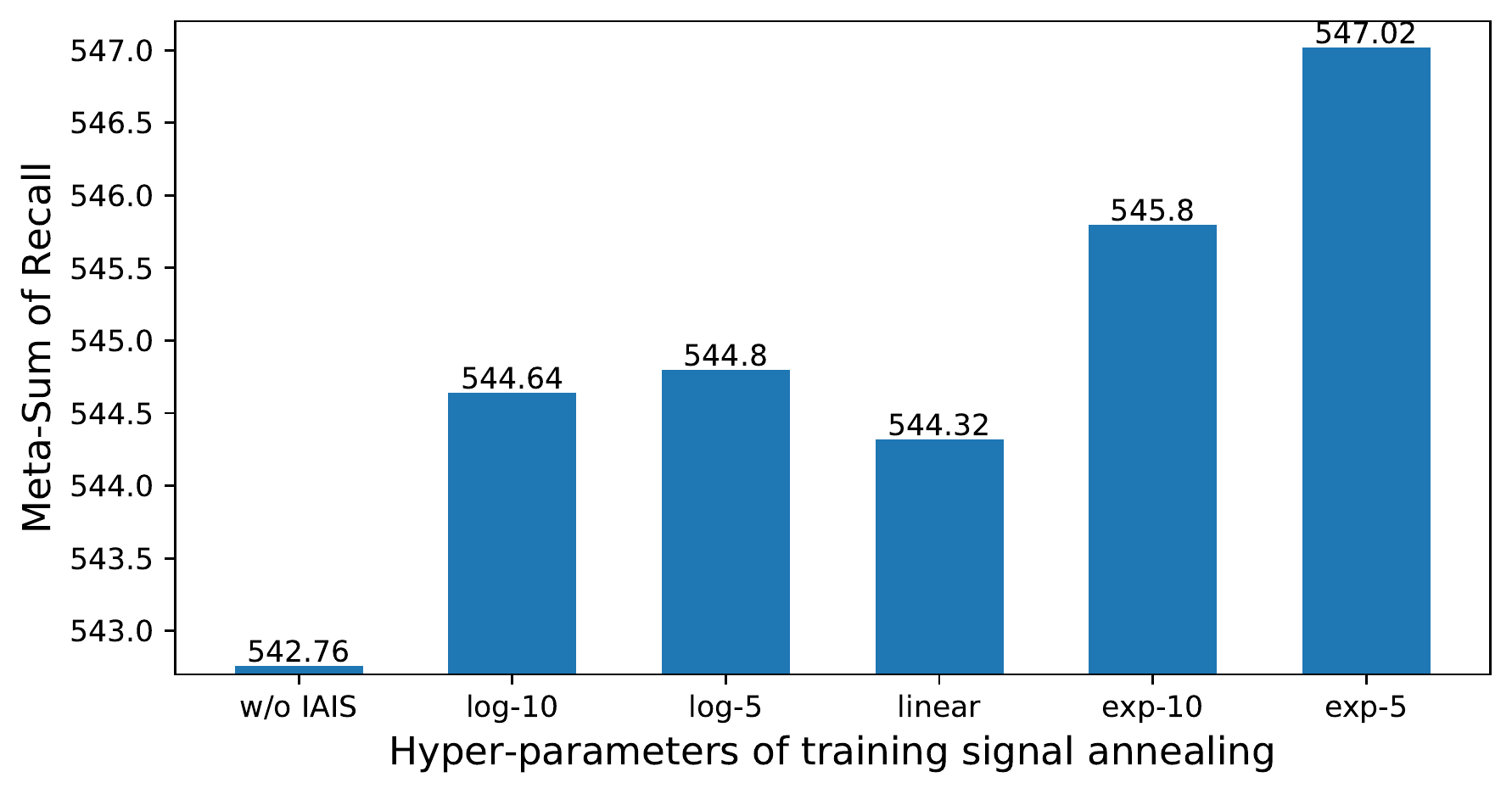}
    \caption{Comparison of the different hyper-parameters in training signal annealing. The exp schedule with scale $\gamma=5$ achieves the best performance.}
    \label{fig:tsa-tuning}
\end{figure}

We compare the results of five schedules for IAIS signal annealing. 
The results in Figure~\ref{fig:tsa-tuning} show that the exp schedule with scale $\gamma=5$ achieves the best performance.
\subsection{Effect of Layer to Apply IAIS}
\begin{figure}[t!]
    \centering
    \includegraphics[width=0.9\linewidth]{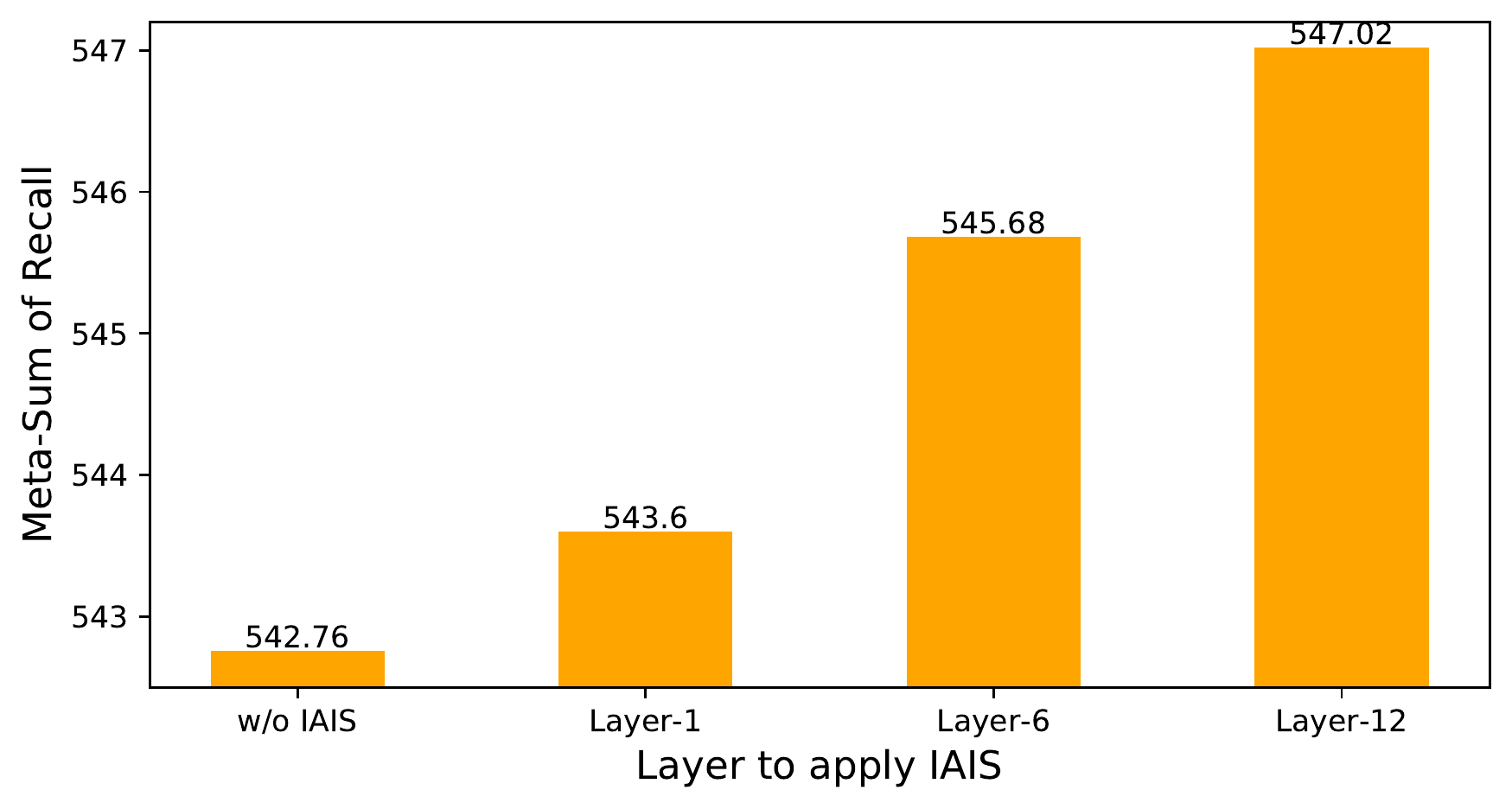}
    \caption{Comparison of the different layers to apply IAIS on the Flickr30k dataset. The IAIS applied on the last layer achieves the best performance.}
    \label{fig:layer_tuning}
\end{figure}

We also apply IAIS on different layers of UNITER-base. 
As illustrated in Figure~\ref{fig:layer_tuning}, the optimal way is to apply IAIS on the last layer.
We speculate that it is more important to learn relation alignment in the deeper layers because the attention in the deeper layers has a bigger impact on the final output, while the effect of the attention in shallow layers might fade away due to the normalization.

\begin{figure*}[t!]
    \centering
    \includegraphics[width=\linewidth]{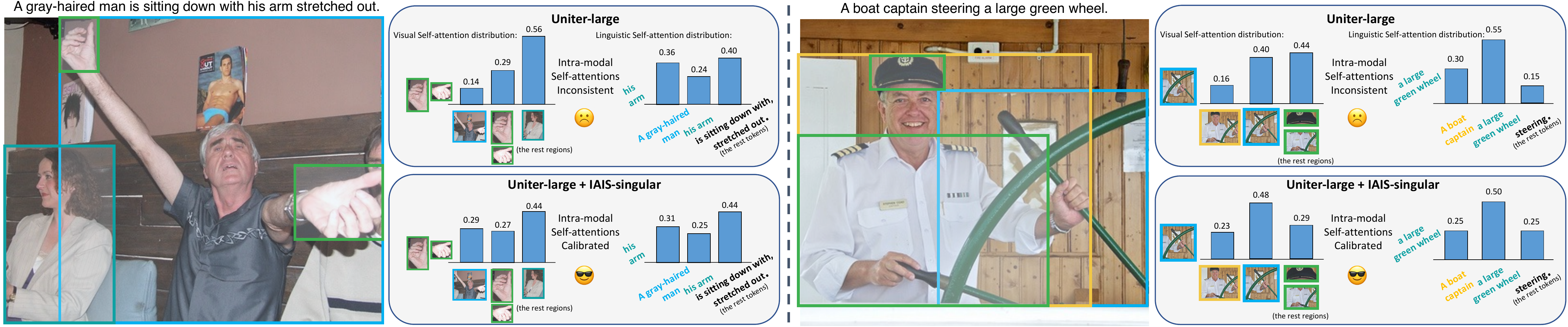}
    \caption{Visualization of intra-modal self-attentions with and without our IAIS method.}
    \label{fig:case-study}
\end{figure*}

\subsection{Case Study}
We further discuss the advantage of our proposed relation-level alignment. 
Figure~\ref{fig:case-study} shows two visualization examples of the intra-modal self-attentions from the Flickr30k Entities dataset. 
With IAIS regularization, the model is instructed to concentrate on the common relations within the linguistic and visual sequence, yielding more calibrated and consistent self-attention distributions. 

\section{Related Work}
In this section, we introduce the task of image-text retrieval and review the representative studies of large-scale multimodal pre-trained models.

\paragraph{Image-Text Retrieval}
Image-Text Retrieval (ITR,~\citealp{Matching-Words-and-Pictures, Learning-the-Semantics-of-Words-and-Pictures}), also known as Image-Text Matching, is one of the popular and challenging Language-and-Vision (V+L) tasks. 
Given image-text pairs, the prevailing approaches project them into a joint representation space, on which cosine or dot-product similarities are defined, and recall the most relevant one according to the similarity. 

\paragraph{Multimodal Pre-trained Models}
The development of the transformer-based large-scale pre-training paradigm sweeps across the area of multimodal learning and achieves many state-of-the-art results on V+L tasks like Image Captioning, Visual Question Answering, Visual Commonsense Reasoning, etc. Recent prevailing multimodal pre-trained models can be categorized into single-stream~\cite{UNITER, VILLA, Inter-BERT, Oscar, VL-BERT, m6} and two-stream~\cite{ERNIE-ViL, LXMERT, ViLBERT} models. Given a piece of text and an image, the former architecture concatenates the features of tokens and regions and learns their joint representations with one transformer model, while the latter embeds the textual and the visual input separately with two independent intra-modal transformers and then utilizes an inter-modal transformer to reinforce cross-modal interactions via cross-modal attention modules. 

\section{Conclusion}
In this paper, we promote the semantic alignment for cross-modal retrieval from the object level to the relation level.
We propose a surrogate metric to quantify the relation consistency by measuring the semantic distance between linguistic and visual relations. 
Furthermore, we present a regularized training method IAIS to calibrate intra-modal self-attentions mutually by minimizing the ISD metric. 
Our method improves both the performance and the interpretability of large-scale pre-trained models. 
Note that, without object annotation in practice, the singular and distributed version of the IAIS loss only provides a coarse-grained attention distribution alignment. 
We leave the elaborate design of ISDa proxy function for future work.

\section*{Acknowledgments}
This work is partly supported by Beijing Academy  of Artificial Intelligence (BAAI). 
We thank all the anonymous reviewers for their constructive comments, and Xuancheng Ren and Lei Li for their helpful suggestions in preparing the manuscript. 

\bibliographystyle{acl_natbib}
\bibliography{anthology,acl2021}

\begin{thebibliography}{33}
\expandafter\ifx\csname natexlab\endcsname\relax\def\natexlab#1{#1}\fi

\bibitem[{Anderson et~al.(2018)Anderson, He, Buehler, Teney, Johnson, Gould,
  and Zhang}]{BUTD}
Peter Anderson, Xiaodong He, Chris Buehler, Damien Teney, Mark Johnson, Stephen
  Gould, and Lei Zhang. 2018.
\newblock \href {https://doi.org/10.1109/CVPR.2018.00636} {Bottom-up and
  top-down attention for image captioning and visual question answering}.
\newblock In \emph{2018 {IEEE} Conference on Computer Vision and Pattern
  Recognition, {CVPR} 2018, Salt Lake City, UT, USA, June 18-22, 2018}, pages
  6077--6086. {IEEE} Computer Society.

\bibitem[{Barnard et~al.(2003)Barnard, Duygulu, Forsyth, de~Freitas, Blei, and
  Jordan}]{Matching-Words-and-Pictures}
Kobus Barnard, Pinar Duygulu, David~A. Forsyth, Nando de~Freitas, David~M.
  Blei, and Michael~I. Jordan. 2003.
\newblock \href {http://jmlr.org/papers/v3/barnard03a.html} {Matching words and
  pictures}.
\newblock \emph{J. Mach. Learn. Res.}, 3:1107--1135.

\bibitem[{Barnard and
  Forsyth(2001)}]{Learning-the-Semantics-of-Words-and-Pictures}
Kobus Barnard and David~A. Forsyth. 2001.
\newblock \href {https://doi.org/10.1109/ICCV.2001.937654} {Learning the
  semantics of words and pictures}.
\newblock In \emph{Proceedings of the Eighth International Conference On
  Computer Vision (ICCV-01), Vancouver, British Columbia, Canada, July 7-14,
  2001 - Volume 2}, pages 408--415. {IEEE} Computer Society.

\bibitem[{Bugliarello et~al.(2020)Bugliarello, Cotterell, Okazaki, and
  Elliott}]{Multimodal-Pretraining-Unmasked}
Emanuele Bugliarello, Ryan Cotterell, Naoaki Okazaki, and Desmond Elliott.
  2020.
\newblock \href {http://arxiv.org/abs/2011.15124} {Multimodal pretraining
  unmasked: Unifying the vision and language berts}.
\newblock \emph{CoRR}, abs/2011.15124.

\bibitem[{Chen et~al.(2020)Chen, Li, Yu, Kholy, Ahmed, Gan, Cheng, and
  Liu}]{UNITER}
Yen{-}Chun Chen, Linjie Li, Licheng Yu, Ahmed~El Kholy, Faisal Ahmed, Zhe Gan,
  Yu~Cheng, and Jingjing Liu. 2020.
\newblock \href {https://doi.org/10.1007/978-3-030-58577-8\_7} {{UNITER:}
  universal image-text representation learning}.
\newblock In \emph{Computer Vision - {ECCV} 2020 - 16th European Conference,
  Glasgow, UK, August 23-28, 2020, Proceedings, Part {XXX}}, volume 12375 of
  \emph{Lecture Notes in Computer Science}, pages 104--120. Springer.

\bibitem[{Clark et~al.(2019)Clark, Khandelwal, Levy, and
  Manning}]{What-Does-BERT-Look-At}
Kevin Clark, Urvashi Khandelwal, Omer Levy, and Christopher~D. Manning. 2019.
\newblock \href {http://arxiv.org/abs/1906.04341} {What does {BERT} look at? an
  analysis of bert's attention}.
\newblock \emph{CoRR}, abs/1906.04341.

\bibitem[{Faghri et~al.(2018)Faghri, Fleet, Kiros, and Fidler}]{VSE++}
Fartash Faghri, David~J. Fleet, Jamie~Ryan Kiros, and Sanja Fidler. 2018.
\newblock \href {http://bmvc2018.org/contents/papers/0344.pdf} {{VSE++:}
  improving visual-semantic embeddings with hard negatives}.
\newblock In \emph{British Machine Vision Conference 2018, {BMVC} 2018,
  Newcastle, UK, September 3-6, 2018}, page~12. {BMVA} Press.

\bibitem[{Gan et~al.(2020)Gan, Chen, Li, Zhu, Cheng, and Liu}]{VILLA}
Zhe Gan, Yen{-}Chun Chen, Linjie Li, Chen Zhu, Yu~Cheng, and Jingjing Liu.
  2020.
\newblock \href
  {https://proceedings.neurips.cc/paper/2020/hash/49562478de4c54fafd4ec46fdb297de5-Abstract.html}
  {Large-scale adversarial training for vision-and-language representation
  learning}.
\newblock In \emph{Advances in Neural Information Processing Systems 33: Annual
  Conference on Neural Information Processing Systems 2020, NeurIPS 2020,
  December 6-12, 2020, virtual}.

\bibitem[{Htut et~al.(2019)Htut, Phang, Bordia, and
  Bowman}]{Do-Attention-Heads-Track-Syntactic-Dependencies}
Phu~Mon Htut, Jason Phang, Shikha Bordia, and Samuel~R. Bowman. 2019.
\newblock \href {http://arxiv.org/abs/1911.12246} {Do attention heads in {BERT}
  track syntactic dependencies?}
\newblock \emph{CoRR}, abs/1911.12246.

\bibitem[{Karpathy and Li(2015)}]{Deep-visual-semantic-alignments}
Andrej Karpathy and Fei{-}Fei Li. 2015.
\newblock \href {https://doi.org/10.1109/CVPR.2015.7298932} {Deep
  visual-semantic alignments for generating image descriptions}.
\newblock In \emph{{IEEE} Conference on Computer Vision and Pattern
  Recognition, {CVPR} 2015, Boston, MA, USA, June 7-12, 2015}, pages
  3128--3137. {IEEE} Computer Society.

\bibitem[{Kiros et~al.(2014)Kiros, Salakhutdinov, and
  Zemel}]{Unifying-Visual-Semantic}
Ryan Kiros, Ruslan Salakhutdinov, and Richard~S. Zemel. 2014.
\newblock \href {http://arxiv.org/abs/1411.2539} {Unifying visual-semantic
  embeddings with multimodal neural language models}.
\newblock \emph{CoRR}, abs/1411.2539.

\bibitem[{Kovaleva et~al.(2019)Kovaleva, Romanov, Rogers, and
  Rumshisky}]{Revealing-the-Dark-Secrets-of-BERT}
Olga Kovaleva, Alexey Romanov, Anna Rogers, and Anna Rumshisky. 2019.
\newblock \href {https://doi.org/10.18653/v1/D19-1445} {Revealing the dark
  secrets of {BERT}}.
\newblock In \emph{Proceedings of the 2019 Conference on Empirical Methods in
  Natural Language Processing and the 9th International Joint Conference on
  Natural Language Processing, {EMNLP-IJCNLP} 2019, Hong Kong, China, November
  3-7, 2019}, pages 4364--4373. Association for Computational Linguistics.

\bibitem[{Lee et~al.(2018)Lee, Chen, Hua, Hu, and He}]{SCAN}
Kuang{-}Huei Lee, Xi~Chen, Gang Hua, Houdong Hu, and Xiaodong He. 2018.
\newblock \href {https://doi.org/10.1007/978-3-030-01225-0\_13} {Stacked cross
  attention for image-text matching}.
\newblock In \emph{Computer Vision - {ECCV} 2018 - 15th European Conference,
  Munich, Germany, September 8-14, 2018, Proceedings, Part {IV}}, volume 11208
  of \emph{Lecture Notes in Computer Science}, pages 212--228. Springer.

\bibitem[{Li et~al.(2020)Li, Yin, Li, Zhang, Hu, Zhang, Wang, Hu, Dong, Wei,
  Choi, and Gao}]{Oscar}
Xiujun Li, Xi~Yin, Chunyuan Li, Pengchuan Zhang, Xiaowei Hu, Lei Zhang, Lijuan
  Wang, Houdong Hu, Li~Dong, Furu Wei, Yejin Choi, and Jianfeng Gao. 2020.
\newblock \href {https://doi.org/10.1007/978-3-030-58577-8\_8} {Oscar:
  Object-semantics aligned pre-training for vision-language tasks}.
\newblock In \emph{Computer Vision - {ECCV} 2020 - 16th European Conference,
  Glasgow, UK, August 23-28, 2020, Proceedings, Part {XXX}}, volume 12375 of
  \emph{Lecture Notes in Computer Science}, pages 121--137. Springer.

\bibitem[{Lin et~al.(2021)Lin, Men, Yang, Zhou, Ding, Zhang, Wang, Wang, Jiang,
  Jia, Zhang, Zhang, Zou, Li, Deng, Liu, Xue, Zhou, Ma, Yu, Li, Lin, Zhou,
  Tang, and Yang}]{m6}
Junyang Lin, Rui Men, An~Yang, Chang Zhou, Ming Ding, Yichang Zhang, Peng Wang,
  Ang Wang, Le~Jiang, Xianyan Jia, Jie Zhang, Jianwei Zhang, Xu~Zou, Zhikang
  Li, Xiaodong Deng, Jie Liu, Jinbao Xue, Huiling Zhou, Jianxin Ma, Jin Yu,
  Yong Li, Wei Lin, Jingren Zhou, Jie Tang, and Hongxia Yang. 2021.
\newblock \href {http://arxiv.org/abs/2103.00823} {{M6:} {A} chinese multimodal
  pretrainer}.
\newblock \emph{CoRR}, abs/2103.00823.

\bibitem[{Lin et~al.(2020)Lin, Yang, Zhang, Liu, Zhou, and Yang}]{Inter-BERT}
Junyang Lin, An~Yang, Yichang Zhang, Jie Liu, Jingren Zhou, and Hongxia Yang.
  2020.
\newblock \href {http://arxiv.org/abs/2003.13198} {Interbert:
  Vision-and-language interaction for multi-modal pretraining}.
\newblock \emph{CoRR}, abs/2003.13198.

\bibitem[{Lin et~al.(2014)Lin, Maire, Belongie, Hays, Perona, Ramanan,
  Doll{\'{a}}r, and Zitnick}]{COCO}
Tsung{-}Yi Lin, Michael Maire, Serge~J. Belongie, James Hays, Pietro Perona,
  Deva Ramanan, Piotr Doll{\'{a}}r, and C.~Lawrence Zitnick. 2014.
\newblock \href {https://doi.org/10.1007/978-3-319-10602-1\_48} {Microsoft
  {COCO:} common objects in context}.
\newblock In \emph{Computer Vision - {ECCV} 2014 - 13th European Conference,
  Zurich, Switzerland, September 6-12, 2014, Proceedings, Part {V}}, volume
  8693 of \emph{Lecture Notes in Computer Science}, pages 740--755. Springer.

\bibitem[{Liu et~al.(2020)Liu, Mao, Zhang, Xie, Wang, and
  Zhang}]{Graph-Structured-Network}
Chunxiao Liu, Zhendong Mao, Tianzhu Zhang, Hongtao Xie, Bin Wang, and Yongdong
  Zhang. 2020.
\newblock \href {https://doi.org/10.1109/CVPR42600.2020.01093} {Graph
  structured network for image-text matching}.
\newblock In \emph{2020 {IEEE/CVF} Conference on Computer Vision and Pattern
  Recognition, {CVPR} 2020, Seattle, WA, USA, June 13-19, 2020}, pages
  10918--10927. {IEEE}.

\bibitem[{Lu et~al.(2019)Lu, Batra, Parikh, and Lee}]{ViLBERT}
Jiasen Lu, Dhruv Batra, Devi Parikh, and Stefan Lee. 2019.
\newblock \href
  {http://papers.nips.cc/paper/8297-vilbert-pretraining-task-agnostic-visiolinguistic-representations-for-vision-and-language-tasks}
  {Vilbert: Pretraining task-agnostic visiolinguistic representations for
  vision-and-language tasks}.
\newblock In \emph{Advances in Neural Information Processing Systems 32: Annual
  Conference on Neural Information Processing Systems 2019, NeurIPS 2019,
  December 8-14, 2019, Vancouver, BC, Canada}, pages 13--23.

\bibitem[{Messina et~al.(2020)Messina, Amato, Esuli, Falchi, Gennaro, and
  Marchand{-}Maillet}]{TERAN}
Nicola Messina, Giuseppe Amato, Andrea Esuli, Fabrizio Falchi, Claudio Gennaro,
  and St{\'{e}}phane Marchand{-}Maillet. 2020.
\newblock \href {http://arxiv.org/abs/2008.05231} {Fine-grained visual textual
  alignment for cross-modal retrieval using transformer encoders}.
\newblock \emph{CoRR}, abs/2008.05231.

\bibitem[{Plummer et~al.(2015)Plummer, Wang, Cervantes, Caicedo, Hockenmaier,
  and Lazebnik}]{Flickr30k-Entities}
Bryan~A. Plummer, Liwei Wang, Chris~M. Cervantes, Juan~C. Caicedo, Julia
  Hockenmaier, and Svetlana Lazebnik. 2015.
\newblock \href {https://doi.org/10.1109/ICCV.2015.303} {Flickr30k entities:
  Collecting region-to-phrase correspondences for richer image-to-sentence
  models}.
\newblock In \emph{2015 {IEEE} International Conference on Computer Vision,
  {ICCV} 2015, Santiago, Chile, December 7-13, 2015}, pages 2641--2649. {IEEE}
  Computer Society.

\bibitem[{Ren et~al.(2015)Ren, He, Girshick, and Sun}]{faster-rcnn}
Shaoqing Ren, Kaiming He, Ross~B. Girshick, and Jian Sun. 2015.
\newblock \href
  {http://papers.nips.cc/paper/5638-faster-r-cnn-towards-real-time-object-detection-with-region-proposal-networks}
  {Faster {R-CNN:} towards real-time object detection with region proposal
  networks}.
\newblock In \emph{Advances in Neural Information Processing Systems 28: Annual
  Conference on Neural Information Processing Systems 2015, December 7-12,
  2015, Montreal, Quebec, Canada}, pages 91--99.

\bibitem[{Sennrich et~al.(2016)Sennrich, Haddow, and Birch}]{improving-MT}
Rico Sennrich, Barry Haddow, and Alexandra Birch. 2016.
\newblock \href {https://doi.org/10.18653/v1/p16-1009} {Improving neural
  machine translation models with monolingual data}.
\newblock In \emph{Proceedings of the 54th Annual Meeting of the Association
  for Computational Linguistics, {ACL} 2016, August 7-12, 2016, Berlin,
  Germany, Volume 1: Long Papers}. The Association for Computer Linguistics.

\bibitem[{Su et~al.(2020)Su, Zhu, Cao, Li, Lu, Wei, and Dai}]{VL-BERT}
Weijie Su, Xizhou Zhu, Yue Cao, Bin Li, Lewei Lu, Furu Wei, and Jifeng Dai.
  2020.
\newblock \href {https://openreview.net/forum?id=SygXPaEYvH} {{VL-BERT:}
  pre-training of generic visual-linguistic representations}.
\newblock In \emph{8th International Conference on Learning Representations,
  {ICLR} 2020, Addis Ababa, Ethiopia, April 26-30, 2020}. OpenReview.net.

\bibitem[{Tan and Bansal(2019)}]{LXMERT}
Hao Tan and Mohit Bansal. 2019.
\newblock \href {https://doi.org/10.18653/v1/D19-1514} {{LXMERT:} learning
  cross-modality encoder representations from transformers}.
\newblock In \emph{Proceedings of the 2019 Conference on Empirical Methods in
  Natural Language Processing and the 9th International Joint Conference on
  Natural Language Processing, {EMNLP-IJCNLP} 2019, Hong Kong, China, November
  3-7, 2019}, pages 5099--5110. Association for Computational Linguistics.

\bibitem[{Vaswani et~al.(2017)Vaswani, Shazeer, Parmar, Uszkoreit, Jones,
  Gomez, Kaiser, and Polosukhin}]{Attention-is-all-you-need}
Ashish Vaswani, Noam Shazeer, Niki Parmar, Jakob Uszkoreit, Llion Jones,
  Aidan~N. Gomez, Lukasz Kaiser, and Illia Polosukhin. 2017.
\newblock \href {http://papers.nips.cc/paper/7181-attention-is-all-you-need}
  {Attention is all you need}.
\newblock In \emph{Advances in Neural Information Processing Systems 30: Annual
  Conference on Neural Information Processing Systems 2017, December 4-9, 2017,
  Long Beach, CA, {USA}}, pages 5998--6008.

\bibitem[{Wang et~al.(2019)Wang, Liu, Li, Sheng, Yan, Wang, and Shao}]{CAMP}
Zihao Wang, Xihui Liu, Hongsheng Li, Lu~Sheng, Junjie Yan, Xiaogang Wang, and
  Jing Shao. 2019.
\newblock \href {https://doi.org/10.1109/ICCV.2019.00586} {{CAMP:} cross-modal
  adaptive message passing for text-image retrieval}.
\newblock In \emph{2019 {IEEE/CVF} International Conference on Computer Vision,
  {ICCV} 2019, Seoul, Korea (South), October 27 - November 2, 2019}, pages
  5763--5772. {IEEE}.

\bibitem[{Xie et~al.(2020)Xie, Dai, Hovy, Luong, and Le}]{uda}
Qizhe Xie, Zihang Dai, Eduard~H. Hovy, Thang Luong, and Quoc Le. 2020.
\newblock \href
  {https://proceedings.neurips.cc/paper/2020/hash/44feb0096faa8326192570788b38c1d1-Abstract.html}
  {Unsupervised data augmentation for consistency training}.
\newblock In \emph{Advances in Neural Information Processing Systems 33: Annual
  Conference on Neural Information Processing Systems 2020, NeurIPS 2020,
  December 6-12, 2020, virtual}.

\bibitem[{Yang et~al.(2020)Yang, Chen, Zhang, and Sun}]{Visual-Agreement}
Pengcheng Yang, Boxing Chen, Pei Zhang, and Xu~Sun. 2020.
\newblock \href {https://aaai.org/ojs/index.php/AAAI/article/view/6484} {Visual
  agreement regularized training for multi-modal machine translation}.
\newblock In \emph{The Thirty-Fourth {AAAI} Conference on Artificial
  Intelligence, {AAAI} 2020, The Thirty-Second Innovative Applications of
  Artificial Intelligence Conference, {IAAI} 2020, The Tenth {AAAI} Symposium
  on Educational Advances in Artificial Intelligence, {EAAI} 2020, New York,
  NY, USA, February 7-12, 2020}, pages 9418--9425. {AAAI} Press.

\bibitem[{Young et~al.(2014)Young, Lai, Hodosh, and Hockenmaier}]{Flickr30k}
Peter Young, Alice Lai, Micah Hodosh, and Julia Hockenmaier. 2014.
\newblock \href
  {https://tacl2013.cs.columbia.edu/ojs/index.php/tacl/article/view/229} {From
  image descriptions to visual denotations: New similarity metrics for semantic
  inference over event descriptions}.
\newblock \emph{Trans. Assoc. Comput. Linguistics}, 2:67--78.

\bibitem[{Yu et~al.(2020)Yu, Tang, Yin, Sun, Tian, Wu, and Wang}]{ERNIE-ViL}
Fei Yu, Jiji Tang, Weichong Yin, Yu~Sun, Hao Tian, Hua Wu, and Haifeng Wang.
  2020.
\newblock \href {http://arxiv.org/abs/2006.16934} {Ernie-vil: Knowledge
  enhanced vision-language representations through scene graph}.
\newblock \emph{CoRR}, abs/2006.16934.

\bibitem[{Zhang et~al.(2020{\natexlab{a}})Zhang, Hu, Jain, Ie, and
  Sha}]{Denotation-Graph}
Bowen Zhang, Hexiang Hu, Vihan Jain, Eugene Ie, and Fei Sha.
  2020{\natexlab{a}}.
\newblock \href {https://doi.org/10.18653/v1/2020.emnlp-main.60} {Learning to
  represent image and text with denotation graph}.
\newblock In \emph{Proceedings of the 2020 Conference on Empirical Methods in
  Natural Language Processing, {EMNLP} 2020, Online, November 16-20, 2020},
  pages 823--839. Association for Computational Linguistics.

\bibitem[{Zhang et~al.(2020{\natexlab{b}})Zhang, Yin, Ren, Li, and Li}]{DCA}
Zhihan Zhang, Zhiyi Yin, Shuhuai Ren, Xinhang Li, and Shicheng Li.
  2020{\natexlab{b}}.
\newblock \href {https://doi.org/10.1007/978-3-030-60457-8\_1} {{DCA:}
  diversified co-attention towards informative live video commenting}.
\newblock In \emph{Natural Language Processing and Chinese Computing - 9th
  {CCF} International Conference, {NLPCC} 2020, Zhengzhou, China, October
  14-18, 2020, Proceedings, Part {II}}, volume 12431 of \emph{Lecture Notes in
  Computer Science}, pages 3--15. Springer.

\end{thebibliography}


\end{document}